\documentclass[sigconf,nonacm]{acmart}
\usepackage{fontawesome5}

\setcopyright{none}
\acmConference[KDD Cup 2026 UniRec Workshop]
{KDD Cup 2026 Tencent UniRec Challenge Workshop}
{August 12, 2026}
{Jeju, Korea}

\acmYear{2026}
\AtBeginDocument{%
  }

\newcommand\blfootnote[1]{%
  \begingroup
  \renewcommand\thefootnote{}%
  \footnotetext{#1}%
  \addtocounter{footnote}{-1}%
  \endgroup
}

\begin{document}

\title{Field-Aware RankMixer with Dual-Stream Bilinear Fusion for the Tencent UNI-REC Challenge}


\author{Yufeng Zhang}
\authornote{These authors contributed equally to this work.}
\affiliation{%
  \institution{Independent Researcher}
  \city{Shanghai}
  \country{China}}
\email{metazyf@gmail.com}

\author{Zhengqi Xu}
\authornotemark[1]
\affiliation{%
  \institution{Meituan}
  \city{Shanghai}
  \country{China}}
\email{xzq1207105685@gmail.com}

\author{Jiajun Cui}
\authornotemark[1]
\affiliation{%
  \institution{Meituan}
  \city{Shanghai}
  \country{China}}
\email{cuijj96@gmail.com}

\renewcommand{\shortauthors}{Zhang et al.}

\begin{abstract}
This paper presents our solution to the KDD Cup 2026 Tencent UNI-REC Challenge. The task requires joint modeling of multi-domain user behavior sequences and non-sequential multi-field features for target-ad pCVR prediction. We develop a Field-Aware RankMixer (FA-RankMixer) with dual-stream bilinear fusion. The model first applies target-aware DIN modules to extract user interests from multiple behavior domains. It also models recent and earlier interests separately for the longest behavior sequence. The model then forms semantic tokens based on feature fields and behavior domains and uses RankMixer blocks for cross-token interaction. A shallow MLP stream complements the deep RankMixer stream, and a group-wise bilinear module fuses their representations. Our final solution ranks \textbf{ninth} on the official leaderboard. Our code is available at
\href{https://github.com/PixelCookie-zyf/TAAC-2026-SeRankMixer}{%
  \textcolor{blue}{\faGithub\ PixelCookie-zyf/TAAC-2026-SeRankMixer}}.
\end{abstract}


\maketitle
\blfootnote{\scriptsize
KDD Cup 2026 Tencent UniRec Challenge Workshop, August 12, 2026, Jeju, Korea.\\
Competition website: \url{https://algo.qq.com/}.
}

\section{Introduction}
Large-scale advertising recommender systems need to identify user interests from many candidate items in real time. They also need to accurately estimate the predicted conversion rate (pCVR) of a target ad. Prediction quality directly affects user experience and platform business value. Existing recommendation models mainly follow two directions. Sequence models learn dynamic user interests from behavior histories~\cite{zhou2019dien}, while feature interaction models capture high-order relations among user, ad, and context features~\cite{wang2021dcnv2}. However, these two types of information are often processed by separate networks with different structures, limiting their interaction and complicating computation and model scaling~\cite{huang2026mixformer}. Therefore, combining multi-domain behavior sequences and non-sequential multi-field features in a unified and scalable architecture remains an important problem in industrial recommendation.

The Tencent UNI-REC Challenge, jointly held by TAAC 2026 and KDD Cup 2026, focuses on this problem. Participants need to model multi-domain user behavior sequences and non-sequential multi-field features together to predict the pCVR of a target ad. The challenge focuses on prediction accuracy and encourages unified recommendation architectures built with homogeneous and stackable blocks, together with studies on model scaling.

In this work, we develop FA-RankMixer for pCVR prediction. It combines a Field-Aware RankMixer backbone with two-stream bilinear fusion. The model uses the target ad as a query and applies DIN to aggregate behaviors from multiple domains. It separately models recent and earlier interests in the longest behavior sequence. The model then forms semantic tokens by feature field and behavior domain. RankMixer blocks model cross-token interactions, and a group-wise bilinear module fuses a deep RankMixer branch with a shallow MLP branch.

Our main contributions are as follows:
\begingroup
\setlength{\leftmargini}{1.2em}
\begin{itemize}
\setlength{\itemsep}{0pt}
\setlength{\topsep}{2pt}
\item 
We develop FA-RankMixer, which combines field and domain-aware semantic tokenization, target-aware multi-domain DIN, RankMixer blocks, and group-wise bilinear fusion for unified feature modeling.
\item Our final solution ranks ninth on the official leaderboard of the TAAC-2026 Tencent UNI-REC Challenge, showing the practical value of FA-RankMixer for large-scale pCVR prediction.
\end{itemize}
\endgroup

\section{Related Work}
Large recommendation models mainly follow two directions: building scalable Mixer architectures and jointly modeling behavior sequences with non-sequential features.

\subsection{Scalable Mixer Architectures}
Some recent work focuses on scalable Mixer architectures for feature interaction. RankMixer~\cite{zhu2025rankmixer} combines parameter-free token mixing with per-token feed-forward networks. TokenMixer-Large \cite{jiang2026tokenmixer} improves model depth and scale through stronger residual paths and Sparse Per-token MoE, while UniMixer~\cite{ha2026unimixer} replaces fixed token mixing with a learnable parameterized module. These methods mainly process feature tokens or sequence summaries rather than raw behavior sequences.

\subsection{Unified Sequence and Feature Modeling}
Another direction combines behavior sequences with other feature fields. InterFormer~\cite{zeng2025interformer} exchanges information through global, sequence, and bridging branches. OneTrans~\cite{zhang2026onetrans} uses one causal Transformer stream, while MixFormer~\cite{huang2026mixformer} co-scales dense features and behavior sequences in one backbone. HyFormer~\cite{huang2026hyformer} alternates Query Decoding and Query Boosting, and TokenFormer~\cite{zhou2026tokenformer} treats fields, behaviors, and targets as one entity stream. Our method instead follows a staged design. It first extracts target-aware representations from each behavior domain and then combines them with multi-field features through RankMixer and dual-stream bilinear fusion. The next section describes these stages in detail.

\section{Methodology}
\begin{figure}[t]
  \centering
  \includegraphics[width=0.9\linewidth]{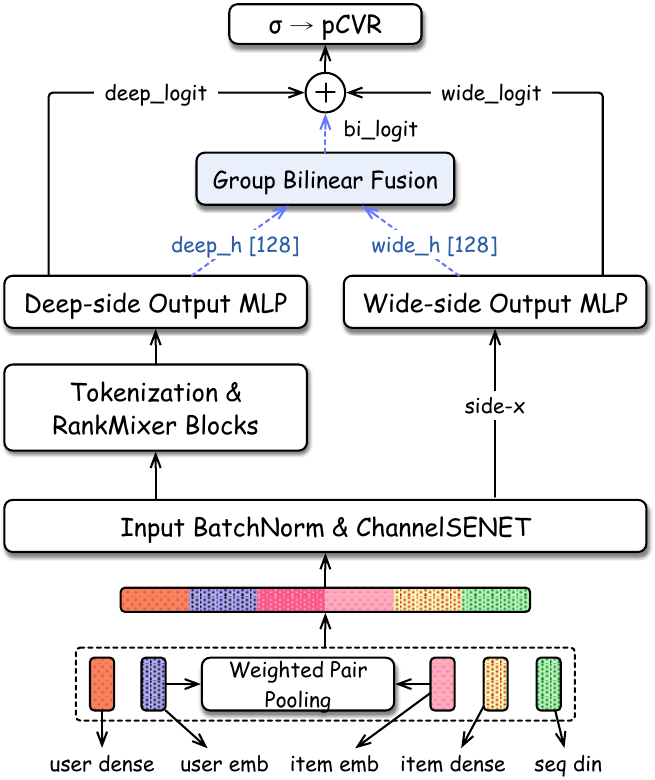}
  \caption{Overview of FA-RankMixer. Target-aware sequence representations and
  non-sequential fields are concatenated and reweighted before entering the
  deep RankMixer stream and the shallow MLP stream. Group-wise bilinear fusion
  combines the two streams for final prediction.}
  \Description{User and item fields and multi-domain sequence representations
  enter weighted pair pooling, input normalization, and ChannelSENET. Parallel
  RankMixer and shallow MLP streams process the result before group-wise
  bilinear fusion and sigmoid pCVR prediction.}
  \label{fig:framework}
\end{figure}

This section presents FA-RankMixer for pCVR prediction. We first formulate the task and then describe its three main components: target-aware feature construction, the Field-Aware RankMixer backbone, and dual-stream bilinear fusion. Figure~\ref{fig:framework} shows the overall architecture. The model constructs target-aware representations from sparse fields, dense fields, and multi-domain behavior sequences. After input normalization and channel reweighting, a deep RankMixer stream and a shallow MLP stream process the same feature vector. Their logits and hidden interactions are combined to predict pCVR.

\subsection{Problem Formulation}
For each click sample, the input contains user features $\mathbf{x}^{u}$, target
ad features $\mathbf{x}^{i}$, and behavior sequences
$\{\mathcal{S}^{m}\}_{m\in\mathcal{M}}$ from multiple domains. Each feature set
contains sparse and dense fields. A sequence
$\mathcal{S}^{m}=\{(\mathbf{s}^{m}_{j},\Delta t^{m}_{j})\}_{j=1}^{L_m}$ records
the behavior features and their time gaps from the current click. Given a
binary conversion label $y\in\{0,1\}$, the task is to estimate
\begin{equation}
  \hat{y}=P\!\left(y=1\mid\mathbf{x}^{u},\mathbf{x}^{i},
  \{\mathcal{S}^{m}\}_{m\in\mathcal{M}}\right)=\sigma(z),
  \label{eq:task}
\end{equation}
where $z$ is the output logit and $\sigma(\cdot)$ is the sigmoid function.

\subsection{Target-Aware Feature Construction}
Non-sequential fields describe relatively stable user and ad attributes, while behavior sequences reflect dynamic interests related to the target ad. Since these two sources have different structures and meanings, processing them in the same way may lose useful information. We therefore construct two complementary representations: field representations for non-sequential features and target-aware representations for multi-domain sequences. They are then concatenated and reweighted to form the shared input for the RankMixer and shallow MLP streams.

\begin{figure}[t]
  \centering
  \includegraphics[width=\linewidth]{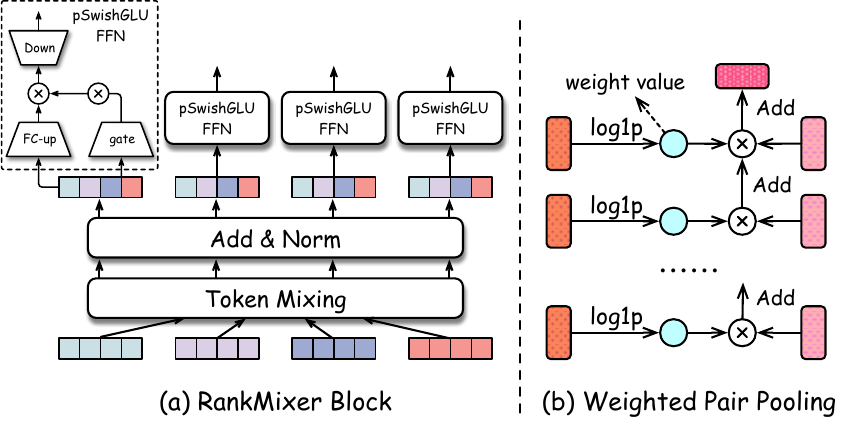}
  \caption{Structures of two key components. (a) a RankMixer block with
  parameter-free token mixing and token-specific pSwiGLU FFNs; (b) weighted
  pair pooling using transformed dense values to weight paired sparse
  embeddings.}
  \Description{The left panel shows token mixing, residual normalization, and
  token-specific pSwiGLU feed-forward networks in a RankMixer block. The right
  panel shows transformed dense values weighting paired sparse embeddings
  before pooling.}
  \label{fig:components}
\end{figure}

\textbf{Non-sequential fields.}
Each sparse field has its own embedding table. For a variable-length sparse field paired with non-negative dense values, we use a weighted mean instead of uniform pooling. The paired value indicates the strength of each ID, which would be ignored if all valid IDs received the same weight. The right side of Figure~\ref{fig:components} illustrates this process:
\begin{equation}
  \mathbf{e}_{f}=
  \frac{\sum_{j}m_{f,j}w_{f,j}E_f(v_{f,j})}
       {\sum_{j}m_{f,j}w_{f,j}+\epsilon},
  \label{eq:weighted-pool}
\end{equation}
where $m_{f,j}$ masks padding and $w_{f,j}$ is the paired dense value. Other multi-value fields use mean pooling. We take the mean of the target-ad field embeddings as the DIN query $\mathbf{q}$.

\textbf{Multi-domain behavior sequences.}
We next use the target ad to aggregate behaviors from each domain. The embeddings of all event fields are concatenated and projected to a $d$-dimensional event vector. A bucketed time-gap embedding is then added to
retain temporal information. Let $\mathbf{e}^{m}_{j}$ denote the resulting
vector. DIN scores each event using the target ad:
\begin{align}
  a^{m}_{j} &= g_m\!\left([\mathbf{q},\mathbf{e}^{m}_{j},
  \mathbf{q}-\mathbf{e}^{m}_{j},
  \mathbf{q}\odot\mathbf{e}^{m}_{j}]\right), \\
  \alpha^{m}_{j} &= \operatorname{softmax}_{j}(a^{m}_{j}), \\
  \mathbf{h}^{m}&=\mathbf{q}+W_o^m\sum_j\alpha^{m}_{j}W_v^m\mathbf{e}^{m}_{j}.
  \label{eq:din}
\end{align}
Padding events are excluded from the softmax. A single DIN over a long sequence can be dominated by recent events and weaken signals from older preferences. Therefore, for the longest domain sequence, the newest-first list is split into
recent and earlier halves. Two independent DIN modules aggregate the halves, and their outputs are preserved by concatenation:
\begin{equation}
  \mathbf{h}^{m^*}=
  [\mathbf{h}^{m^*}_{\mathrm{recent}}\,\|\,
   \mathbf{h}^{m^*}_{\mathrm{earlier}}].
  \label{eq:dual-din}
\end{equation}
Finally, user embeddings, target-ad embeddings, sequence representations, and dense fields are concatenated. These blocks have different value ranges, so batch normalization first aligns their scales. ChannelSENET~\cite{hu2018squeeze} then reweights informative channels, $\widetilde{\mathbf{x}}=\mathbf{x}\odot \sigma(g_{\mathrm{se}}(\mathbf{x}))$. The resulting post-SENET vector is shared by the RankMixer and shallow MLP streams.

\subsection{Field-Aware RankMixer}
We next convert the shared post-SENET vector into semantic tokens and model
their interactions with RankMixer~\cite{zhu2025rankmixer}. Mapping all dense
features to a single token would hide their field identities. We instead assign
each dense field, or a small predefined group of related dense fields, to a
separate tokenizer group. User and target-ad sparse blocks and each behavior
domain also use separate projections. For group $g$, the tokenizer maps the
corresponding slice $\widetilde{\mathbf{x}}_g$ to $n_g$ tokens:
\begin{equation}
  \mathbf{X}_g=\operatorname{reshape}
  (W_g\widetilde{\mathbf{x}}_g+\mathbf{b}_g,n_g,D),\quad
  \mathbf{X}=[\mathbf{X}_1\|\cdots\|\mathbf{X}_G].
  \label{eq:tokenizer}
\end{equation}
Our final model uses $T=24$ tokens with dimension $D=1152$: five user tokens,
three target-ad tokens, four behavior-domain tokens, nine user-dense tokens,
and three item-dense tokens. The left side of Figure~\ref{fig:components} shows
the structure of a RankMixer block.

Each RankMixer block retains parameter-free token mixing. We split every token
into $T$ equal parts. The $h$-th output token concatenates the $h$-th part from
all input tokens:
\begin{equation}
  \operatorname{Mix}(\mathbf{X})_h=
  [\mathbf{x}^{(h)}_1\|\mathbf{x}^{(h)}_2\|\cdots\|
  \mathbf{x}^{(h)}_T], \quad h=1,\ldots,T.
  \label{eq:token-mixing}
\end{equation}
This operation exchanges information across tokens without introducing mixing parameters. We replace LayerNorm in the Original RankMixer blocks with RMSNorm to avoid changing the original offset information of features due to mean centering. Then Given the input $\mathbf{X}^{\ell}$, the $\ell$-th RankMixer block is computed as follows:
\begin{align}
  \mathbf{S}^{\ell} &= \operatorname{RMSNorm}
  (\operatorname{Mix}(\mathbf{X}^{\ell})+\mathbf{X}^{\ell}), \\
  \mathbf{X}^{\ell+1} &= \operatorname{RMSNorm}
  (\operatorname{PFFN}(\mathbf{S}^{\ell})+\mathbf{S}^{\ell}).
  \label{eq:rankmixer}
\end{align}
We also replace the standard per-token FFN with token-specific pSwiGLU~\cite{jiang2026tokenmixer}. Tokens from different feature fields and behavior domains have different meanings. Therefore, each token uses its own pSwiGLU parameters instead of sharing one FFN across all tokens. The learned gate controls which dimensions of the transformed token are passed to the output. For token $t$, we have
\begin{equation}
  \operatorname{PFFN}_t(\mathbf{s}_t)=
  \left(\operatorname{SiLU}(\mathbf{s}_tW^{t}_{\mathrm{gate}})
  \odot(\mathbf{s}_tW^{t}_{\mathrm{up}})\right)W^{t}_{\mathrm{down}}.
  \label{eq:pswiglu}
\end{equation}
The matrices are not shared across tokens. We stack two blocks and mean-pool the final tokens to obtain the deep representation, which is then passed to the deep prediction MLP.

\subsection{Dual-Stream Bilinear Fusion and Objective}
To combine high-order token interactions with a direct feature path, we use a deep RankMixer stream and a shallow MLP stream. The deep stream applies an MLP with hidden sizes 1024, 512, 256, and 128 to the RankMixer output. In parallel, the shallow stream maps the post-SENET feature vector through hidden sizes 512 and 128. Let
$\mathbf{h}^{d},\mathbf{h}^{w}\in\mathbb{R}^{128}$ be their last hidden vectors, and let $z_d$ and $z_w$ be their logits. We divide both hidden vectors into $K$ groups and compute as follows:
\begin{equation}
  z_{\mathrm{bi}}=\sum_{k=1}^{K}
  (\mathbf{h}^{d}_{k})^{\top}W_k\mathbf{h}^{w}_{k},\qquad
  z=z_d+z_w+z_{\mathrm{bi}}.
  \label{eq:bilinear}
\end{equation}
The bilinear matrices are initialized to zero, so the interaction starts as a residual over the two-stream logits. We observe that the positive conversion labels in the dataset are sparse, so an unweighted objective can be dominated by negative samples. We therefore use positive-class weighted binary cross-entropy to increase the gradient contribution of converted samples:
\begin{equation}
  \mathcal{L}=-\frac{1}{N}\sum_{n=1}^{N}
  \left[\lambda y_n\log\hat{y}_n+(1-y_n)\log(1-\hat{y}_n)\right],
  \label{eq:loss}
\end{equation}
where $\lambda=2.0$ in our final setting.

\section{Experiments}
This section evaluates FA-RankMixer on the TAAC-2026 Tencent UNI-REC Challenge. We first describe the experimental setup, then analyze the effects of the main components and model scaling and report the overall performance.

\subsection{Experimental Setup}
\noindent\textbf{Dataset and evaluation.}
We use the Industrial Track data from the second-round TAAC-2026 Tencent UNI-REC Challenge for post-click conversion rate prediction. The training set contains 34,822,423 labeled clicks over 10 consecutive days, and the official test set contains 12,251,082 clicks from the following two days. The positive rate is 7.8867\%. The observed conversion delay has a 99th percentile of 66.823 hours, and our audit finds no evidence of systematic false negatives caused by delayed feedback. Both the official leaderboard and our local validation use ROC-AUC.

\noindent\textbf{Validation protocol.}
We reserve the last 10\% of Parquet row groups for development. This is not a temporal split: the row groups are not sorted by timestamp, and most groups span almost the full 10-day window. The resulting development set is therefore close to a random sample over the training period. We do not apply another time-based split and select the training epoch by development AUC.

\noindent\textbf{Model configuration.}
The final model uses 24 semantic tokens: five user tokens, three target-ad tokens, four behavior-domain tokens, and twelve dense-field tokens. RankMixer has width 1152, 24 heads, two blocks, pSwiGLU expansion 2, RMSNorm, and mean pooling. Both the embedding and sequence widths are 64. The four sequence limits are 256, 256, 512, and 1024; two DIN modules aggregate the recent and earlier halves of the longest sequence. The 7,399-dimensional concatenated input passes through a $[1024,256]$ ChannelSENET. The deep and shallow MLPs use $[1024,512,256,128]$ and $[512,128]$, respectively, followed by eight-group bilinear fusion. The dense backbone has about 425.6M parameters, and the total is about 601M under the vocabulary used for counting. Fields with more than one million IDs do not use embedding tables.

\noindent\textbf{Optimization.}
We use weighted binary cross-entropy with a positive weight of 2.0. Muon~\cite{liu2025muon} optimizes matrix parameters with learning rate $5\times10^{-4}$, momentum 0.95, weight decay 0.01, and five Newton--Schulz steps. AdamW~\cite{loshchilov2019decoupled} with learning rate $1\times10^{-4}$ handles the remaining dense parameters, while Adagrad~\cite{duchi2011adaptive} with learning rate 0.05 updates embeddings. We use a constant learning rate and SAM~\cite{foret2021sharpness} with $\rho=0.05$. We apply SWA~\cite{izmailov2018averaging} to dense weights starting at epoch 2 and collect snapshots every 200 steps and at epoch boundaries. BatchNorm statistics are recomputed with 50 batches. Other settings are dropout 0.01, 6 GPUs, batch size 1024 per GPU, FP32 training, and $\operatorname{log1p}$ transformation for selected long-tailed dense fields.

\subsection{Ablation Analysis}

\begin{figure*}[t]
  \centering
  \includegraphics[width=\textwidth]{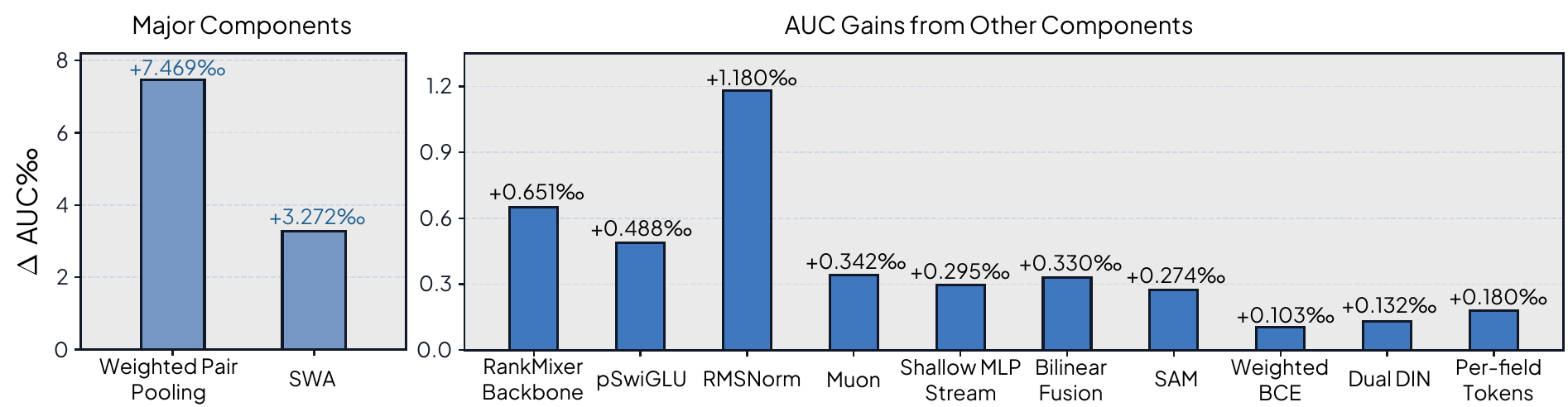}
  \caption{Incremental Test AUC gains from components added to FA-RankMixer.
  The left panel highlights the two largest gains, while the right panel shows
  gains from the remaining components using a different vertical scale.}
  \Description{The left panel shows gains of 7.469 per mille from Weighted Pair
  Pooling and 3.272 per mille from SWA. The right panel shows ten smaller
  positive gains from the remaining components on a different vertical scale.}
  \label{fig:ablation}
\end{figure*}

We conduct an incremental ablation study to quantify the contribution of each component. Starting from the baseline, each configuration adds one component to the preceding configuration, and all configurations are evaluated by Test AUC on the official test set. As shown in Figure~\ref{fig:ablation}, Weighted Pair Pooling and SWA are the two dominant contributors, improving Test AUC by $7.469\text{\textperthousand}$ and $3.272\text{\textperthousand}$, respectively. The substantial gain from Weighted Pair Pooling confirms that the paired dense values provide informative strength signals when aggregating multi-value sparse fields. The improvement from SWA suggests that averaging dense weights further enhances the generalization of the final model.

The remaining components provide smaller but consistent improvements. RMSNorm delivers the largest gain among them, while refinements to the backbone, token-specific FFNs, optimization, dual-stream fusion, loss weighting, and sequence modeling all contribute positively. The final transition to per-field dense tokenization with a width of 1152 further improves performance. These results indicate that the final gain arises from a coordinated combination of architectural and optimization choices rather than a single modification. Overall, FA-RankMixer improves Test AUC from 0.814098 to 0.828814, corresponding to a total gain of $14.716\text{\textperthousand}$.

\subsection{Scaling Analysis}

\begin{figure}[t]
  \centering
  \includegraphics[width=\linewidth]{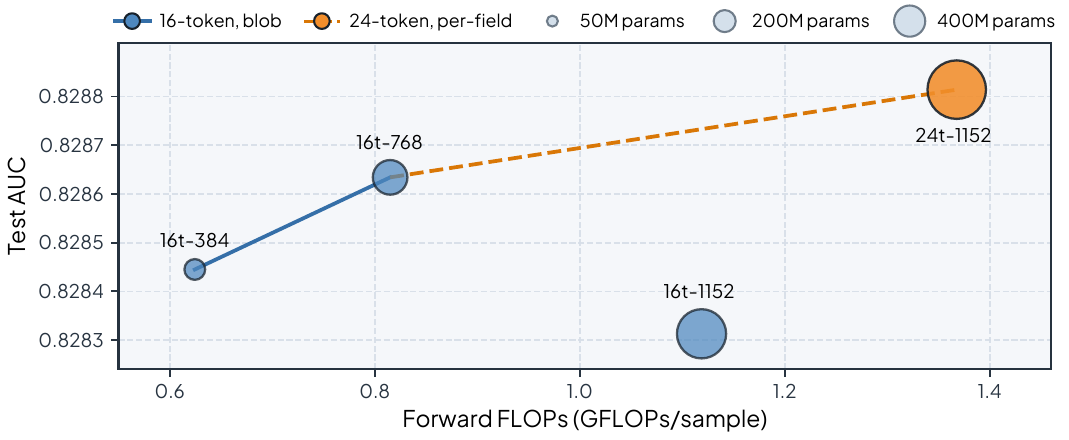}
  \caption{Compute--accuracy scaling of FA-RankMixer. Bubble area represents the dense parameter count. Solid and dashed lines denote width-only scaling and the transition to per-field tokenization, respectively.}
  \Description{A bubble plot compares forward GFLOPs per sample with Test AUC
  for four configurations. Bubble area indicates dense parameter count. A
  solid blue line connects the 16-token width-384 and width-768 models, a
  dashed orange line connects the width-768 model to the final 24-token
  per-field model, and the degraded 16-token width-1152 model is unconnected.}
  \label{fig:scaling}
\end{figure}

We further investigate how FA-RankMixer scales with model width and token granularity. Figure~\ref{fig:scaling} summarizes the resulting compute--accuracy trade-off. Forward FLOPs are measured per sample for a single inference pass using PyTorch FlopCounterMode, counting two FLOPs per multiply--accumulate; embedding lookups are excluded. Under the fixed 16-token blob tokenization, increasing the width from 384 to 768 improves Test AUC, whereas further widening the model to 1152 degrades performance despite the additional parameters and computation. This result indicates that width-only scaling does not yield monotonic gains. We therefore jointly increase the token capacity and replace coarse blob tokenization with 24 per-field tokens, allowing the model to preserve field identities at a finer semantic granularity. The resulting configuration achieves the best Test AUC of 0.828814 at 1.368 GFLOPs per sample. Overall, effective scaling depends not only on model size, but also on how capacity is organized across semantic tokens.

\subsection{Overall Performance}
Our final 24-token per-field FA-RankMixer achieves a Test AUC of 0.828814 and ranks ninth on the official leaderboard of the TAAC-2026 Tencent UNI-REC Challenge. This result demonstrates the practical effectiveness of the proposed architecture for large-scale industrial pCVR prediction.

\section{Conclusion and Future Work}
In this paper, we present FA-RankMixer for unified modeling of multi-domain behavior sequences and non-sequential multi-field features. It combines target-aware interest extraction, field-aware semantic tokenization, RankMixer blocks, and dual-stream bilinear fusion. The final model achieves a Test AUC of 0.828814 and ranks ninth. Our leaderboard result and post-competition experiments provide two practical insights: performance comes from coordinated architectural and optimization choices, and finer-grained field-aware tokens are more effective than width-only scaling.

Our evidence is limited to a single competition dataset, while the 601M-parameter model has substantial deployment cost. To support reproducibility, we release our implementation and report the main experimental settings. Future work should investigate adaptive field-aware tokenization that assigns computation according to the information content of each field, together with distillation and sparse computation for efficient inference.
\bibliographystyle{ACM-Reference-Format}
\bibliography{sample-base}

@String{Computer = "{IEEE} Computer" }

@inproceedings{zhu2025rankmixer,
  title = {RankMixer: Scaling Up Ranking Models in Industrial Recommenders},
  author = {Zhu, Jie and Fan, Zhifang and Zhu, Xiaoxie and Jiang, Yuchen and Wang, Hangyu and Han, Xintian and Ding, Haoran and Wang, Xinmin and Zhao, Wenlin and Gong, Zhen and others},
  booktitle = {Proceedings of the 34th ACM International Conference on Information and Knowledge Management},
  pages = {6309--6316},
  year = {2025},
  address = {Seoul, Republic of Korea}
}

@article{jiang2026tokenmixer,
  title = {TokenMixer-Large: Scaling Up Large Ranking Models in Industrial Recommenders},
  author = {Jiang, Yuchen and Zhu, Jie and Han, Xintian and Lu, Hui and Bai, Kunmin and Yang, Mingyu and Wu, Shikang and Zhang, Ruihao and Zhao, Wenlin and Bai, Shipeng and others},
  journal = {arXiv preprint arXiv:2602.06563},
  year = {2026}
}

@article{ha2026unimixer,
  title = {UniMixer: A Unified Architecture for Scaling Laws in Recommendation Systems},
  author = {Ha, Mingming and Wang, Guanchen and Chen, Linxun and Rao, Xuan and Shi, Yuexin and Ma, Tianbao and Liu, Zhaojie and Fan, Yunqian and Lu, Zilong and Niu, Yanan and others},
  journal = {arXiv preprint arXiv:2604.00590},
  year = {2026}
}

@inproceedings{zeng2025interformer,
  title = {InterFormer: Effective Heterogeneous Interaction Learning for Click-Through Rate Prediction},
  author = {Zeng, Zhichen and Liu, Xiaolong and Hang, Mengyue and Liu, Xiaoyi and Zhou, Qinghai and Yang, Chaofei and Liu, Yiqun and Ruan, Yichen and Chen, Laming and Chen, Yuxin and others},
  booktitle = {Proceedings of the 34th ACM International Conference on Information and Knowledge Management},
  pages = {6225--6233},
  year = {2025},
  address = {Seoul, Republic of Korea}
}

@inproceedings{zhang2026onetrans,
  title = {OneTrans: Unified Feature Interaction and Sequence Modeling with One Transformer in Industrial Recommender},
  author = {Zhang, Zhaoqi and Pei, Haolei and Guo, Jun and Wang, Tianyu and Feng, Yufei and Sun, Hui and Liu, Shaowei and Sun, Aixin},
  booktitle = {Proceedings of the ACM Web Conference 2026},
  pages = {8162--8170},
  year = {2026},
  address = {Dubai, United Arab Emirates}
}

@article{huang2026mixformer,
  title = {MixFormer: Co-Scaling Up Dense and Sequence in Industrial Recommenders},
  author = {Huang, Xu and Zhang, Hao and Fan, Zhifang and Huang, Yunwen and Wei, Zhuoxing and Chai, Zheng and Ni, Jinan and Zheng, Yuchao and Chen, Qiwei},
  journal = {arXiv preprint arXiv:2602.14110},
  year = {2026}
}

@article{huang2026hyformer,
  title = {HyFormer: Revisiting the Roles of Sequence Modeling and Feature Interaction in CTR Prediction},
  author = {Huang, Yunwen and Hong, Shiyong and Xiao, Xijun and Jin, Jinqiu and Luo, Xuanyuan and Wang, Zhe and Chai, Zheng and Wu, Shikang and Zheng, Yuchao and Lin, Jingjian},
  journal = {arXiv preprint arXiv:2601.12681},
  year = {2026}
}

@article{zhou2026tokenformer,
  title = {TokenFormer: Unify the Multi-Field and Sequential Recommendation Worlds},
  author = {Zhou, Yifeng and Hu, Yuehong and Feng, Zhixiang and Pan, Junwei and Wu, Kaihui and Li, Hanyong and Zhang, Shangyu and Huang, Shudong and Zhu, Zhangbin and Yin, Chengguo and others},
  journal = {arXiv preprint arXiv:2604.13737},
  year = {2026}
}

@article{liu2025muon,
  title = {{Muon} is Scalable for {LLM} Training},
  author = {Liu, Jingyuan and Su, Jianlin and Yao, Xingcheng and Jiang, Zhejun and Lai, Guokun and Du, Yulun and Qin, Yidao and Xu, Weixin and Lu, Enzhe and Yan, Junjie and others},
  journal = {arXiv preprint arXiv:2502.16982},
  year = {2025}
}

@inproceedings{loshchilov2019decoupled,
  title = {Decoupled Weight Decay Regularization},
  author = {Loshchilov, Ilya and Hutter, Frank},
  booktitle = {7th International Conference on Learning Representations},
  address = {New Orleans, LA, USA},
  year = {2019}
}

@inproceedings{foret2021sharpness,
  title = {Sharpness-Aware Minimization for Efficiently Improving Generalization},
  author = {Foret, Pierre and Kleiner, Ariel and Mobahi, Hossein and Neyshabur, Behnam},
  booktitle = {9th International Conference on Learning Representations},
  address = {Virtual Event, Austria},
  year = {2021}
}

@article{duchi2011adaptive,
  title = {Adaptive Subgradient Methods for Online Learning and Stochastic Optimization},
  author = {Duchi, John and Hazan, Elad and Singer, Yoram},
  journal = {Journal of Machine Learning Research},
  volume = {12},
  pages = {2121--2159},
  year = {2011}
}

@inproceedings{izmailov2018averaging,
  title = {Averaging Weights Leads to Wider Optima and Better Generalization},
  author = {Izmailov, Pavel and Podoprikhin, Dmitrii and Garipov, Timur and Vetrov, Dmitry and Wilson, Andrew Gordon},
  booktitle = {Proceedings of the Thirty-Fourth Conference on Uncertainty in Artificial Intelligence},
  pages = {876--885},
  year = {2018}
}

@inproceedings{hu2018squeeze,
  title = {Squeeze-and-Excitation Networks},
  author = {Hu, Jie and Shen, Li and Sun, Gang},
  booktitle = {Proceedings of the IEEE Conference on Computer Vision and Pattern Recognition},
  pages = {7132--7141},
  address = {Salt Lake City, UT, USA},
  year = {2018}
}

@inproceedings{zhou2019dien,
  title={Deep interest evolution network for click-through rate prediction},
  author={Zhou, Guorui and Mou, Na and Fan, Ying and Pi, Qi and Bian, Weijie and Zhou, Chang and Zhu, Xiaoqiang and Gai, Kun},
  booktitle={Proceedings of the AAAI conference on artificial intelligence},
  volume={33},
  number={01},
  pages={5941--5948},
  year={2019},
  address={Honolulu, Hawaii, USA}
}

@inproceedings{wang2021dcnv2,
  title={Dcn v2: Improved deep \& cross network and practical lessons for web-scale learning to rank systems},
  author={Wang, Ruoxi and Shivanna, Rakesh and Cheng, Derek and Jain, Sagar and Lin, Dong and Hong, Lichan and Chi, Ed},
  booktitle={Proceedings of the web conference 2021},
  pages={1785--1797},
  year={2021},
  address={Ljubljana, Slovenia}
}


\end{document}